\documentclass[
]{ceurart}

\usepackage{listings}
\usepackage[table]{xcolor}
\usepackage{booktabs}
\usepackage{wrapfig}
\usepackage{tikz}
\usetikzlibrary{positioning}
\begin{document}

\copyrightyear{2026}
\copyrightclause{Copyright for this paper by its authors.
  Use permitted under Creative Commons License Attribution 4.0
  International (CC BY 4.0).}

\conference{SLM4ED'26: The 1st Workshop of Small Language Models for Education (SLM4ED),
  June 28, 2026, Seoul, Republic of Korea}

\title{CSTutorBench: Benchmarking Small Language Models as Tutors for Block-Based Programming}

\author[1]{H. Chad Lane}
\author[1]{Bryson Kageler}
\address[1]{University of Illinois Urbana-Champaign, Champaign, IL, USA}

\begin{abstract}
Large language models are increasingly explored as AI tutors, yet deploying them in K–12 settings raises concerns around privacy, cost, and reliance on proprietary models. Small language models (SLMs) offer a promising alternative, but selecting the right model for a specific educational context remains difficult, particularly when the target domain, such as block-based programming, is largely absent from model training data. We introduce \textit{CSTutorBench}, a benchmark for evaluating language models as CS tutors in VEX VR, a block-based robotics environment. The benchmark comprises 17 scenario-based questions scored against a pedagogical rubric grounded in established tutoring and feedback research, with a human-in-the-loop LLM-as-judge pipeline for evaluation. Preliminary findings across 11 models (4B–120B parameters) reveal that models perform well on surface-level criteria such as vocabulary and tone but struggle with deeper pedagogical behaviors, particularly avoiding answer leakage and engaging with student debugging histories. In our sample, model family and instruction-tuning approach appear to be better predictors of tutoring quality than parameter count alone, though the small number of models limits the strength of this conclusion. A targeted prompt revision grounded in recent educational prompt engineering research improved scores for 10 of 11 models. These results underscore the value of context-specific, pedagogically grounded benchmarks for SLM selection in educational deployment.
\end{abstract}


\maketitle

\section{Introduction}

Large language models (LLMs) have demonstrated considerable potential as educational tools, particularly for tasks that involve generating explanations, providing feedback, and engaging learners in dialogue \cite{shi2026large}. For computer science education, it is particularly appealing that many models are trained on large code corpora, meaning they can reliably produce syntactically correct solutions, trace through program logic, and explain programming concepts with impressive fluency \cite{jiang2026survey}. But, echoing a classic lesson learned in AI in Education showing that it is unwise to simply wrap a tutor around an expert system \cite{clancey1984Neomycin}: we should not expect an LLM that is good at coding to be good at \emph{teaching} people how to code. Effective tutoring involves a delicate balance of providing guidance, but allowing learners to maintain a sense of agency \cite{chi2001learning, vanlehn2011relative, wood2012role}. Although some benchmarks do focus on the educational capabilities of LLMs \cite{xu2025edubench} and others on tutoring \cite{srinivasa2025tutorbench}, there continues to be a need to more directly assess the tutoring and coaching capabilities of modern LLMs and investigate new methods for customizing their behavior.

These challenges are amplified when the target population consists of young learners working in block-based programming environments such as Scratch, VEX, or Snap!. Block-based environments are designed specifically for education and are largely absent from the training data of even the most capable models. At the same time, practical constraints in K-12 settings around privacy, limited budgets, and the need for a high degree of local control, all point toward small language models (SLMs) as a viable path for addressing these needs. But SLMs vary widely in their behavior and there is little guidance to deciding which 8B or 30B model to deploy as a classroom tutor. In this paper, we introduce \textit{CSTutorBench}, a benchmark for evaluating language models as CS tutors in VEX VR, a block-based robotics simulation for middle school students. The benchmark includes 17 scenario-based questions, an 8-criterion pedagogical rubric, and an automated LLM-as-judge evaluation pipeline. We present preliminary results comparing models from 4B to 120B parameters and discuss implications for SLM selection in educational contexts.

\section{CSTutorBench}

\begin{wrapfigure}{r}{0.525\textwidth}
  \centering
  \vspace{-12pt}
  \includegraphics[width=0.495\textwidth]{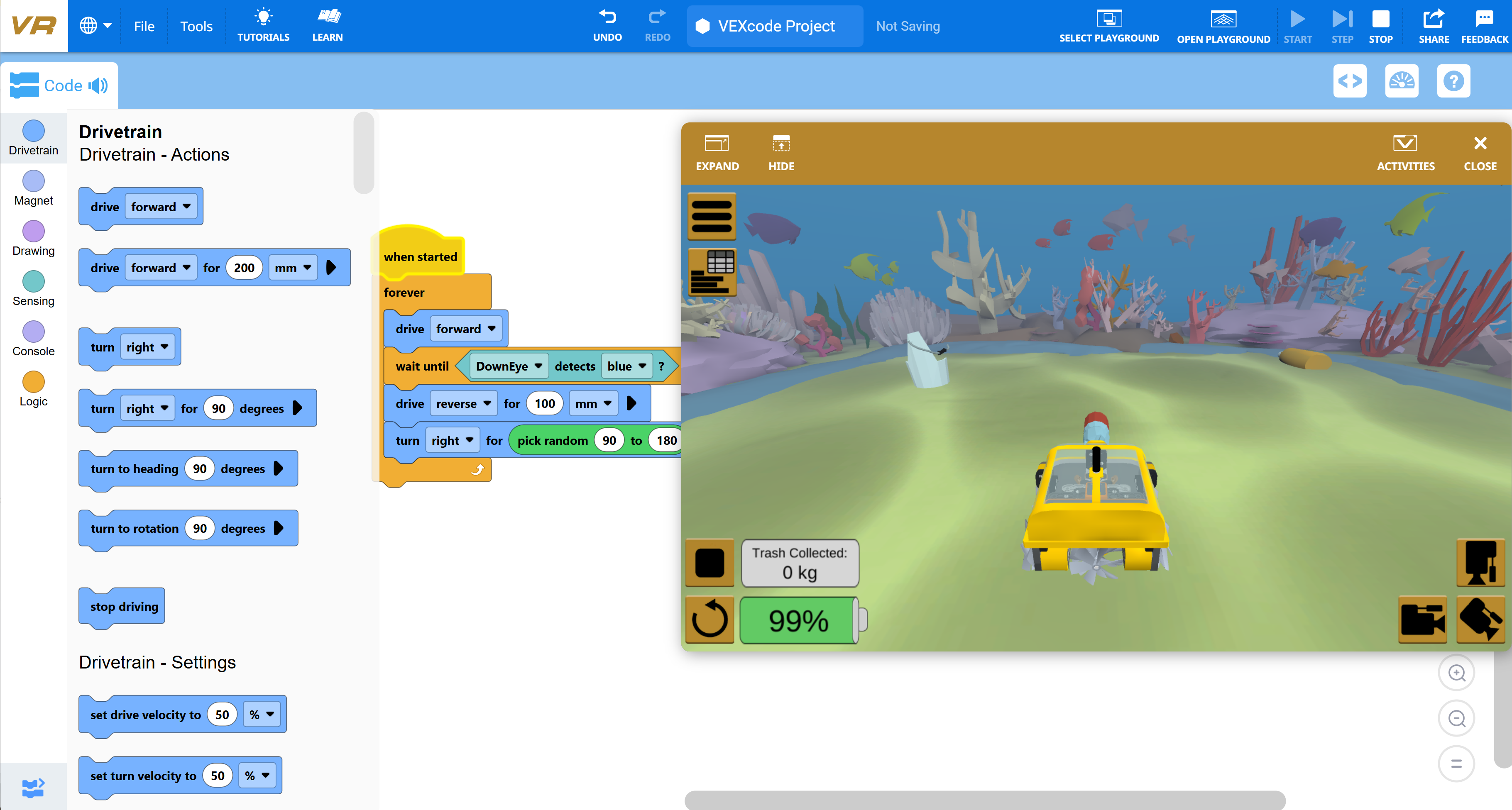}
  \caption{VEX VR Coral Reef Rescue environment.}
  \label{fig:coralreef}
  \vspace{-10pt}
\end{wrapfigure}

\paragraph{Task Domain.}
CSTutorBench is deliberately scoped to a single task and audience: middle school students (grades 6--8) working on VEX VR Coral Reef Rescue, a simulation in which students program a virtual robot to navigate an underwater environment and collect trash using a drag-and-drop Blockly interface. The task spans three levels of increasing complexity: Level~1 introduces sequencing and basic movement, Level~2 adds distance sensor navigation, and Level~3 requires algorithmic optimization. We needed a benchmark that draws from the actual task, audience, and pedagogical goals. General-purpose benchmarks cannot capture domain-specific requirements such as knowing that a VEX \texttt{Drive Forward} block is non-blocking or that the simulator arena is 2000\,mm $\times$ 2000\,mm. Student programs are represented as Blockly XML, a format largely absent from standard code
training corpora, making this a particularly revealing test of whether a model can tutor effectively in a domain where it lacks deep pre-training.

\paragraph{Dataset.}
The benchmark contains 17 scenario-based questions across four types: \textit{debugging}~(8), \textit{debugging\_iterative}~(6), \textit{optimization}~(1), and \textit{conceptual}~(2). The questions are not drawn from real student data but are inspired by common errors and bugs we observe students making when working with VEX~VR, including missing forever loops, inverted conditional logic, blocking commands, sensor misuse, and confusion between absolute and relative headings. Each entry includes the student's question text, one or more Blockly XML code snapshots, an 8-criterion rubric with universal and entry-specific scoring guides, and evaluator notes describing the correct diagnosis (hidden from the model under test).

The six \textit{debugging\_iterative} entries are designed to approximate multi-turn tutoring interactions. Each entry outlines several turns of student effort: 2--3 code snapshots showing the student's self-directed attempts at fixing a bug before asking for help. For example, one entry presents a student who first tries a \texttt{repeat-10} loop, then switches to \texttt{repeat-100}, trying to understand why their robot keeps stopping. The model receives the full iteration history and must engage with the student's progression rather than treating it as a fresh question. A dedicated rubric criterion, \textit{acknowledges\_progression}, specifically evaluates whether the response recognizes and builds on these prior attempts.

\paragraph{Rubric.}
Each entry is scored on eight criteria (0--2 points each, 16 points max): seven universal criteria applied to every question plus one type-specific criterion that varies by question type. The rubric operationalizes established research on tutoring and formative feedback into computable dimensions. Table~\ref{tab:rubric} summarizes each criterion, what it measures, and its grounding in the literature.

\begin{table}[t]
\centering
\caption{CSTutorBench rubric criteria. Each criterion is scored 0 (no credit), 1 (partial), or 2 (full credit). Codes in parentheses correspond to column headers in Table~\ref{tab:heatmap}.}
\label{tab:rubric}
\footnotesize
\begin{tabular}{@{}p{3.6cm}p{5.8cm}p{2.7cm}@{}}
\toprule
\textbf{Criterion} & \textbf{What It Measures} & \textbf{Grounding} \\
\midrule
\multicolumn{3}{@{}l}{\textit{Universal --- scored on all 17 questions}} \\[2pt]
Conciseness (Conc) & Response is brief (1--3 sentences) with no padding or restating the question & \cite{shute2008focus,narciss2013designing} \\[3pt]
Vocabulary (Voc) & Language is accessible to a middle schooler; no unexplained jargon & \cite{wiggins2012seven,graesser1995collaborative} \\[3pt]
Accuracy (Acc) & No incorrect claims about block behavior, sensors, or robot physics & \cite{shute2008focus,vanlehn2011relative,black1998assessment} \\[3pt]
Formatting (Fmt) & No raw XML or inappropriate markdown; reads as natural conversation & \cite{shute2008focus,narciss2013designing} \\[3pt]
Tone & Encouraging and patient; appropriate for a middle school student & \cite{lepper2002wisdom,mueller1998praise,hattie2007power} \\[3pt]
Actionability (Act) & Student knows what to look at or try next after reading the response; next-step guidance & \cite{hattie2007power,wiggins2012seven,black1998assessment} \\[3pt]
Targetedness (Tgt) & Engages with this student's specific situation rather than giving generic advice & \cite{lepper2002wisdom,nicol2006formative,vanlehn2011relative} \\[3pt]
\midrule
\multicolumn{3}{@{}l}{\textit{Type-specific --- one per question, determined by question type}} \\[2pt]
Hint not solution (HnS) & Guides the student toward the answer without giving it away ($n$\,=\,8 debugging questions) & \cite{narciss2013designing,shute2008focus} \\[3pt]
Acknowledges progression (AckP) & Recognizes and builds on the student's prior attempts ($n$\,=\,6 iterative debugging questions) & \cite{vanlehn2011relative,narciss2013designing} \\[3pt]
Builds on success (BoS) & Validates correct work rather than hallucinating additional problems ($n$\,=\,1 optimization question) & \cite{narciss2013designing,shute2008focus} \\[3pt]
Conceptual clarity (CC) & Explains block functionality clearly when a direct answer is appropriate ($n$\,=\,2 conceptual questions) & \cite{shute2008focus,vanlehn2011relative} \\
\bottomrule
\end{tabular}
\end{table}

A few criteria stand out that are particularly challenging for LLM-based tutors. \textit{Conciseness} reflects Shute's~\cite{shute2008focus} finding that feedback complexity is inversely related to error correction and learning efficiency: when feedback is too long or too complicated, learners simply stop attending to it. \textit{Actionability} draws on Hattie and Timperley's~\cite{hattie2007power} model of effective feedback as information that reduces the gap between current performance and a desired goal by specifying what to do next. \textit{Targetedness} is grounded in research on expert human tutors~\cite{lepper2002wisdom, vanlehn2011relative}, who consistently engage with the specific contextual factors rather than offering generic advice. VanLehn's~\cite{vanlehn2011relative} meta-analysis further shows that step-level feedback outperforms final-answer feedback, reinforcing the importance of engaging with student work at a fine grain. Wiggins~\cite{wiggins2012seven} argues that effective feedback must be accessible and user-friendly, which motivates the \textit{vocabulary} criterion. \textit{Tone} reflects work on the role of encouragement in expert tutoring~\cite{lepper2002wisdom} and the distinction Mueller and Dweck~\cite{mueller1998praise} draw between meaningful effort-focused praise and empty ability praise. The four type-specific criteria reflect Narciss's~\cite{narciss2013designing} interactive tutoring feedback (ITF) model, which holds that no single feedback type is optimal across all tasks and that feedback must be calibrated to both the learning task and the student's state. Also, recent work by Liu et al.~\cite{liu2025comparative} evaluates LLM and SLM feedback quality along related dimensions including readability, specificity, and actionability, finding that AI feedback excels at clarity and structure while humans provide more contextually nuanced guidance.

Each criterion in Table \ref{tab:rubric} includes a description that is stable across entries and, where needed, an entry-specific addendum describing what correct and incorrect responses look like for that particular question. This structure allows the rubric to maintain consistency across the benchmark while capturing the domain knowledge required for accurate scoring.

\paragraph{Evaluation Pipeline.}
We evaluate 11 models spanning six families (Google Gemma, Alibaba Qwen, NVIDIA Nemotron, DeepSeek, Allen AI OLMo, and OpenAI GPT) with parameter counts ranging from 4B to 120B. Each model is tested under two system prompt conditions. The first prompt provides the VEX VR block reference, assignment description, and instructions to use pedagogical policies, the only specific guideline being not to give away answers, along with constraints such as never referencing raw XML and keeping responses to 1-3 sentences. The second prompt retains these elements but incorporates targeted revisions informed by research on prompt engineering for educational AI \cite{holmes2026prompt,ahn2026mentors}: additional domain-specific knowledge drawn from observed model weaknesses (e.g., clarifying which sensors are irrelevant to the task, specifying blocking behavior of movement commands) and explicit pedagogical scaffolding guidelines (e.g., acknowledge what is working before addressing errors, guide toward discovery rather than stating fixes, recognize student progress across attempts).

Responses are collected via Ollama running locally or through the Google AI Studio API. Scoring follows a human-in-the-loop LLM-as-judge protocol~\cite{shi2025hitl}. In the automated pass, the judge (Claude Sonnet~4) makes one scoring call per criterion per question, receiving the student question, evaluator notes, the criterion's scoring guide, and the model's response, and returning a score of 0, 1, or 2 with a brief rationale. In the human review pass, a reviewer examines every judgment organized by benchmark question across all models and trials, rather than by model, to avoid systematic bias. 

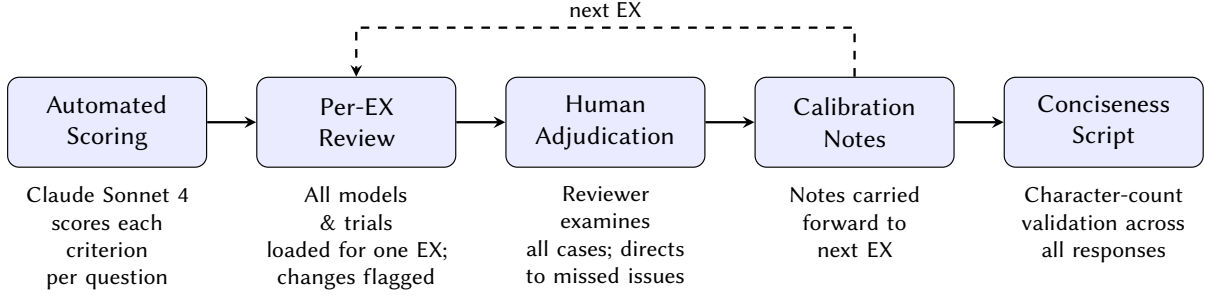
\begin{figure}[t]
\centering
\resizebox{\textwidth}{!}{%
\begin{tikzpicture}[
  node distance=0.6cm,
  box/.style={rectangle, draw, rounded corners, text centered, font=\footnotesize, minimum height=1.1cm, text width=2.2cm, fill=blue!8},
  arrow/.style={->, >=stealth, thick},
  label/.style={font=\scriptsize, text centered, text width=2.2cm}
]
\node[box] (auto) {Automated\\Scoring};
\node[box, right=of auto] (review) {Per-EX\\Review};
\node[box, right=of review] (human) {Human\\Adjudication};
\node[box, right=of human] (calib) {Calibration\\Notes};
\node[box, right=of calib] (conc) {Conciseness\\Script};

\node[label, below=0.1cm of auto] {Claude Sonnet~4\\scores each criterion\\per question};
\node[label, below=0.1cm of review] {All models \& trials\\loaded for one EX;\\changes flagged};
\node[label, below=0.1cm of human] {Reviewer examines\\all cases; directs\\to missed issues};
\node[label, below=0.1cm of calib] {Notes carried\\forward to\\next EX};
\node[label, below=0.1cm of conc] {Character-count\\validation across\\all responses};

\draw[arrow] (auto) -- (review);
\draw[arrow] (review) -- (human);
\draw[arrow] (human) -- (calib);
\draw[arrow] (calib) -- (conc);

\draw[arrow, dashed] (calib.north) -- ++(0,0.6) -| node[above, pos=0.25, font=\scriptsize] {next EX} (review.north);

\end{tikzpicture}
}
\caption{Hybrid evaluation workflow. Each benchmark question (EX) is reviewed across all models and trials before proceeding to the next. Calibration notes accumulate across questions, progressively improving the judge's consistency.}
\label{fig:hybrid}
\end{figure}

For each question, the judge (Sonnet~4) performs a full review of each response and categorizes its suggested changes into confident and borderline cases. The reviewer (human) examines both categories, provides feedback, and directs the judge toward additional issues identified during earlier reviews. Before moving to the next question, a final check is performed, and the judge records calibration notes that carry forward to subsequent questions, progressively improving its consistency (Figure~\ref{fig:hybrid}). For conciseness, the judge proved unreliable at counting sentence equivalents, so a character count script was used to ensure consistent scoring across all responses (each criterion's rubric-defined length thresholds translated to character counts based on 100 characters per sentence). Judge selection is discussed in Section~3. Results are saved incrementally, and the pipeline supports multiple trials per model to assess response variability.

\section{Preliminary Findings}

We evaluated 11~models across six model families, ranging from 4B to 120B parameters (Table~\ref{tab:heatmap}), using the 17~benchmark questions. We tested with two system prompt versions: an initial prompt (Trial~1) and a revised prompt informed by educational prompt engineering patterns from Holmes et~al.~\cite{holmes2026prompt} (Trial~2). All results reported below use the Trial~2 (revised) prompt unless otherwise noted.

\paragraph{Judge Selection and Calibration.}
We initially explored using SLMs as automated judges but found severe leniency bias. Phi4:14b, for example, scored gemma3:27b (Trial~1) responses at 91--96\% across multiple configurations, while a human rater scored the same responses at 61\%. Finding a reliable SLM-based evaluation model remains a priority. We ultimately selected Claude Sonnet~4 as the automated judge based on its closer alignment with the human ground truth, and then iteratively refined the judge's evaluation instructions and rubric across four versions. We validated the final version (v4) against the human-reviewed scores for four models and found that v4 produced overall scores within 0--7 percentage points of the human baseline for three of those four, with one model (deepseek-r1:8b) showing a larger gap. All reported results then underwent the hybrid review process described in Section~2 (Figure~\ref{fig:hybrid}), slightly reducing the gap between automated and human judgments.

\paragraph{Per-Criterion Patterns.}
Table~\ref{tab:heatmap} presents Trial~2 scores broken down by the seven universal rubric criteria plus the four type-specific criteria (each scored on a subset of questions). \textit{Vocabulary} stands out as universally strong, with every model scoring 94\% or above, suggesting that all tested models can reliably produce age-appropriate language for our context. \textit{Tone} is strong for most models (9 of 11 score above 74\%) but shows more variability, with olmo-3 scoring just 50\%.

There is far greater variability on the remaining criteria. \textit{Hint\_not\_solution}, which measures whether a model can guide a student toward a solution without giving it away, is the most challenging dimension. Scores range from 6\% (gpt-oss:20b and olmo-3:7b, which consistently provide direct solutions) to 81\% (gemma-4:31b, gemma3:27b, and nemotron-3-super). \textit{Conciseness} is also highly variable (0--100\%), reflecting a common failure mode where models produce verbose explanations that would overwhelm a middle school student. \textit{Accuracy} on VEX-specific block behavior spans 41--94\%, with lower-performing models hallucinating block functionality or mischaracterizing sensor behavior. Among the type-specific criteria, \textit{acknowledges\_progression} (scored on 6 iterative debugging questions) ranged from 8\% to 67\%, indicating that most models struggle to engage meaningfully with a student's prior attempts even when that history is provided as context.

\begin{table}[t]
  \centering
 \caption{Trial~2 per-criterion scores (\%) evaluated by Claude Sonnet~4 (v4 rubric) with a hybrid review. Models sorted by overall score. Shading:
  \colorbox{blue!20}{$\geq$80\%}\,, \colorbox{yellow!30}{50--79\%}\,, \colorbox{gray!25}{$<$50\%}\,. The first seven criteria are universal (scored
  on all 17 questions). The last four are type-specific, each scored on a subset: HnS on 8, AckP on 6, BoS on 1, CC on 2. Each criterion is scored 0--2 per question; percentages reflect points earned out of the maximum possible (e.g., $17 \times 2 = 34$ for each universal criterion, $8 \times 2 = 16$ for HnS).}

  \label{tab:heatmap}
  \footnotesize
  \setlength{\tabcolsep}{3pt}
  \begin{tabular}{@{}llr r rrrrrrr rrrr@{}}
  \toprule
  & & & & \multicolumn{7}{c}{\textit{Universal}} & \multicolumn{4}{c}{\textit{Type-specific}} \\
  \cmidrule(lr){5-11} \cmidrule(l){12-15}
  \textbf{Model} & \textbf{Family} & \textbf{Size} & \textbf{All} & \textbf{Conc} & \textbf{Voc} & \textbf{Acc} & \textbf{Fmt} & \textbf{Tone} &
  \textbf{Act} & \textbf{Tgt} & \textbf{HnS} & \textbf{AckP} & \textbf{BoS} & \textbf{CC} \\
  \midrule
  gemma-4      & Google   & 31B  & \cellcolor{blue!20}89 & \cellcolor{blue!20}100 & \cellcolor{blue!20}97 & \cellcolor{blue!20}94 &
  \cellcolor{blue!20}100 & \cellcolor{blue!20}94 & \cellcolor{yellow!30}76 & \cellcolor{yellow!30}79 & \cellcolor{blue!20}81 &
  \cellcolor{yellow!30}58 & \cellcolor{yellow!30}50 & \cellcolor{yellow!30}75 \\
  qwen3.5      & Qwen     & 9B   & \cellcolor{yellow!30}79 & \cellcolor{yellow!30}53 & \cellcolor{blue!20}100 & \cellcolor{blue!20}85 &
  \cellcolor{blue!20}85 & \cellcolor{blue!20}88 & \cellcolor{yellow!30}68 & \cellcolor{blue!20}85 & \cellcolor{yellow!30}69 &
  \cellcolor{yellow!30}67 & \cellcolor{blue!20}100 & \cellcolor{yellow!30}75 \\
  qwen3.6      & Qwen     & 35B  & \cellcolor{yellow!30}78 & \cellcolor{gray!25}24 & \cellcolor{blue!20}100 & \cellcolor{blue!20}82 &
  \cellcolor{blue!20}100 & \cellcolor{blue!20}91 & \cellcolor{blue!20}82 & \cellcolor{blue!20}82 & \cellcolor{yellow!30}69 &
  \cellcolor{yellow!30}50 & \cellcolor{gray!25}0 & \cellcolor{yellow!30}75 \\
  deepseek-r1  & DeepSeek & 8B   & \cellcolor{yellow!30}77 & \cellcolor{yellow!30}76 & \cellcolor{blue!20}100 & \cellcolor{yellow!30}76 &
  \cellcolor{blue!20}82 & \cellcolor{blue!20}82 & \cellcolor{yellow!30}62 & \cellcolor{yellow!30}79 & \cellcolor{yellow!30}69 &
  \cellcolor{yellow!30}50 & \cellcolor{gray!25}0 & \cellcolor{yellow!30}75 \\
  nemotron     & NVIDIA   & 120B & \cellcolor{yellow!30}73 & \cellcolor{gray!25}32 & \cellcolor{blue!20}100 & \cellcolor{blue!20}85 &
  \cellcolor{yellow!30}79 & \cellcolor{yellow!30}74 & \cellcolor{blue!20}82 & \cellcolor{yellow!30}74 & \cellcolor{blue!20}81 &
  \cellcolor{gray!25}25 & \cellcolor{gray!25}0 & \cellcolor{yellow!30}75 \\
  gemma4:e4b   & Google   & 4B   & \cellcolor{yellow!30}72 & \cellcolor{gray!25}47 & \cellcolor{blue!20}94 & \cellcolor{yellow!30}71 &
  \cellcolor{blue!20}94 & \cellcolor{blue!20}91 & \cellcolor{yellow!30}59 & \cellcolor{yellow!30}71 & \cellcolor{yellow!30}75 &
  \cellcolor{gray!25}33 & \cellcolor{gray!25}0 & \cellcolor{gray!25}25 \\
  gemma3       & Google   & 27B  & \cellcolor{yellow!30}69 & \cellcolor{gray!25}9 & \cellcolor{blue!20}100 & \cellcolor{yellow!30}76 &
  \cellcolor{yellow!30}76 & \cellcolor{blue!20}91 & \cellcolor{yellow!30}50 & \cellcolor{yellow!30}79 & \cellcolor{blue!20}81 &
  \cellcolor{yellow!30}50 & \cellcolor{yellow!30}50 & \cellcolor{yellow!30}75 \\
  gpt-oss      & OpenAI   & 20B  & \cellcolor{yellow!30}64 & \cellcolor{gray!25}41 & \cellcolor{blue!20}97 & \cellcolor{yellow!30}76 &
  \cellcolor{yellow!30}65 & \cellcolor{yellow!30}71 & \cellcolor{yellow!30}74 & \cellcolor{yellow!30}71 & \cellcolor{gray!25}6 & \cellcolor{gray!25}17
   & \cellcolor{gray!25}0 & \cellcolor{yellow!30}75 \\
  qwen3-coder  & Qwen     & 30B  & \cellcolor{yellow!30}52 & \cellcolor{gray!25}0 & \cellcolor{blue!20}100 & \cellcolor{yellow!30}53 &
  \cellcolor{gray!25}18 & \cellcolor{yellow!30}74 & \cellcolor{yellow!30}53 & \cellcolor{yellow!30}79 & \cellcolor{gray!25}44 & \cellcolor{gray!25}33 &
   \cellcolor{gray!25}0 & \cellcolor{yellow!30}50 \\
  gemma3:4b    & Google   & 4B   & \cellcolor{yellow!30}50 & \cellcolor{gray!25}9 & \cellcolor{blue!20}100 & \cellcolor{gray!25}47 &
  \cellcolor{gray!25}41 & \cellcolor{yellow!30}79 & \cellcolor{gray!25}21 & \cellcolor{yellow!30}74 & \cellcolor{yellow!30}56 & \cellcolor{gray!25}8 &
  \cellcolor{yellow!30}50 & \cellcolor{gray!25}0 \\
  olmo-3       & AI2      & 7B   & \cellcolor{gray!25}49 & \cellcolor{gray!25}41 & \cellcolor{blue!20}100 & \cellcolor{gray!25}41 &
  \cellcolor{yellow!30}56 & \cellcolor{yellow!30}50 & \cellcolor{gray!25}38 & \cellcolor{gray!25}47 & \cellcolor{gray!25}6 & \cellcolor{gray!25}17 &
  \cellcolor{gray!25}0 & \cellcolor{yellow!30}50 \\
  \bottomrule
  \end{tabular}
  \end{table}
  
\paragraph{Model Families and Size.}
The top three models by overall score are gemma-4:31b (89\%), qwen3.5:9b (79\%), and qwen3.6 (78\%). Notably, qwen3.5 achieves this with only 9B parameters, outperforming the 120B nemotron-3-super (73\%) by 6~percentage points. At the other end, the weakest models include olmo-3:7b (49\%) and gemma3:4b (50\%), but even among small models there is wide variation: deepseek-r1 at 8B scores 77\%, well above several larger models, while qwen3-coder at 30B scores just 52\%. With only 11 models spanning six families, we cannot cleanly disentangle the effects of model family, instruction-tuning approach, and parameter count. However, the pattern is suggestive that within our sample, family-level differences appear to influence tutoring quality better than size alone.

\paragraph{Effect of Prompt Revision.}
10~of 11~models improved from Trial~1 to Trial~2, with gains ranging from 6.6 to 16.2 percentage points (mean: 11.2); qwen3-coder:30b was the sole exception, decreasing by 1.5~points. The prompt revision primarily improved the four type-specific criteria along with tone, formatting, and targetedness, while accuracy, actionability, and vocabulary saw minimal change. Conciseness decreased slightly on average, likely because the revised prompt's structured response format encouraged additional scaffolding elements that added length.

\section{Discussion}
The results suggest that both model family and parameter count contribute to tutoring quality, though their effects are difficult to separate in this sample. The Gemma family placed three of its four models in the upper half of the sample, and three of the four models scoring 75\% or higher on \textit{hint\_not\_solution} were Gemma variants, suggesting that something in this family's training recipe supports pedagogically appropriate restraint. The two general-purpose Qwen models also performed well (qwen3.5 at 79\%, qwen3.6 at 78\%), while their code-specialized sibling scored just 52\%. Parameter count alone is clearly insufficient as a predictor: the 120B nemotron-3-super scored 73\%, placing it in the middle of the sample below two models with fewer than 10B parameters, and the 4B gemma4:e4b (72\%) outperformed the 27B gemma3 (69\%), likely reflecting differences in training data choices rather than size. For educators evaluating candidate SLMs, these patterns reinforce the value of targeted evaluation within a specific task context; general-purpose leaderboards and parameter counts are poor proxies for tutoring quality.

We found the poor performance of qwen3-coder:30b to be surprising (just 52\% overall despite being one of the largest model in our sample). This model is explicitly optimized for code generation, yet that specialization appears to have worked against it in a tutoring context. Its formatting score (18\%) and conciseness score (0\%) were the lowest in the sample, suggesting that its training encouraged verbose, developer-oriented output rather than the brief conversational responses appropriate for a middle school student. This result echoes Clancey's~\cite{clancey1984Neomycin} foundational observation that expert-level domain knowledge does not transfer straightforwardly into effective instruction. Even when the domain in question is programming, a model tuned to \emph{produce} code is not necessarily well suited to \emph{teach} someone how to code. For educators selecting SLMs, this suggests that specialist models should be approached with caution, even when their specialty aligns with the subject matter being taught.

A complementary failure mode is visible in gpt-oss:20b, which scored just 6\% on \textit{hint\_not\_solution}. This model is a capable generalist; it scored well on actionability (74\%) and performed adequately on most other criteria. The problem is that it consistently completed productive next steps \emph{for} the student rather than guiding them to do so independently. Even with explicit prompting to withhold answers, this model showed what we might call a helpfulness compulsion: a strong tendency, likely reinforced through RLHF-style training, to resolve the user's problem as directly as possible. This tension between being ``helpful'' in a general-purpose sense and being pedagogically effective is not unique to gpt-oss, but it surfaces most clearly there. The broader pattern across our results, where surface-level criteria are easy and deeper pedagogical criteria are nearly universally failed, may reflect the same underlying tension.

Regarding our system prompt revisions between Trial~1 and Trial~2, we revised it to address weaknesses observed in the initial evaluation. The original prompt (approximately 50 words of instructional content, excluding the injected reference table and assignment description) established a Socratic tutor persona and listed eight rules covering desirable difficulty, code tracing, prediction-based questioning, and focusing on one change at a time. This initial prompt was effective in many ways, but also gave models considerable latitude in how they structured their responses. Many produced lengthy, markdown-formatted output poorly suited to a middle school audience.

The revised prompt drew on two principles from recent work on educational prompt engineering. From Holmes et al.~\cite{holmes2026prompt}, we adopted the combination of the Persona pattern (defining a consistent tutoring role with specific responsibilities) and the Context Manager pattern (explicitly scoping what the model should and should not address), which their evaluation found to be the most effective pairing for educational applications. From Ahn et al.~\cite{ahn2026mentors}, we incorporated elements of the Cognitive Apprenticeship model: graduated mentorship that favors coaching and hinting over direct demonstration (their design guideline DG4), affective behaviors such as affirming student effort before addressing errors (DG5), and grounding feedback in the specific artifact the student has produced (DG6). The revised prompt also added domain-specific context that a real tutoring system would provide: clarifications about sensor behavior in the specific environment, guidance on recognizing when a student's solution is already correct, and notes on common sources of bugs (e.g., confusing continuous and fixed-distance movement blocks). Explicit formatting constraints (no markdown, no bullet points, responses limited to 1--3 sentences) replaced the original prompt's open-ended rules. Overall, the revised prompt roughly quadrupled in length (from approximately 50 to 400 words of instructional content, excluding the injected reference table and assignment description).

10~of 11~models improved from Trial~1 to Trial~2, with gains ranging from 6.6 to 16.2 percentage points (mean gain among the 10 improving models: 11.2); qwen3-coder:30b was the sole exception, decreasing by 1.5~points. At the criterion level, tone showed the largest average gain (18.7\%), followed by the type-specific criteria as a group (17.9\%), formatting, and targetedness, while accuracy, actionability, and vocabulary saw minimal change (4.3\% to 7.0\%). Conciseness showed a small decrease on average (0.8\%), possibly due to the structured response format that the revised system prompt encouraged. Accuracy showed a slight decrease in some models, possibly because the expanded prompt directed more of the model's limited capacity toward scaffolding at the expense of precise VEX-specific claims.

\section{Limitations}

Several limitations of the benchmark design should be noted. The evaluation set currently contains only 17 questions, and while they span four question types and multiple bug categories, this is a small set that limits the precision of the per-criterion comparisons. This was especially true for the type-specific criteria, where \textit{builds\_on\_success} is scored on a single question and \textit{conceptual\_clarity} on just two; percentage scores at this small scale are therefore too sensitive to individual scoring decisions. Additionally, our revised prompt (Trial~2) was developed in response to weaknesses from our first simplified prompt (Trial~1) and tested on the same 17 questions. Given the preliminary status and small size of this test set, we did not hold out a test subset. So the reported gains may conflate genuine improvement with overfitting to the evaluation items. Another limitation is that all items in our benchmark (like most) are single-turn, and thus do not capture critical multi-turn dialogues that characterize actual tutoring. Finally, we have not yet evaluated these models with actual students, and so naturally, higher rubric scores may not actually predict learning outcomes.

The evaluation methodology also has limitations. The automated judge (Claude Sonnet~4) exhibited instancing inconsistency: with no changes to the evaluation instructions, the judge would vary across model-trial combinations in how it weighted or combined criteria, for example merging accuracy and actionability considerations to different degrees. The per-item hybrid review structure was designed to minimize this bias, but the reviewer could not catch every error given the volume of data involved. Additionally, the judge's calibration improved progressively as notes from each question carried forward to subsequent ones, meaning that corrections for earlier benchmark questions may be less thorough than for later ones. Because the review was organized by question across all models rather than by model, this does not introduce systematic bias toward or against any particular model.  Additionally, because the human reviewer examined the judge's scores and rationales before making corrections, anchoring effects may have influenced the final scores. We also report results from a single generation per model per trial, without confidence intervals or variance estimates. Generating multiple runs is straightforward, but each would require the same hybrid review process to produce trustworthy scores. Until a reliable automated evaluator is validated, reporting meaningful run-to-run variance remains impractical at this scale.

\section{Conclusion and Future Work}
Block-based programming environments have proven useful for introductory computer science education tools for younger learners, but also present a challenge for building LLM-driven tutors since coding-focused LLMs are typically trained on industry-grade code. Educationally relevant code has limited representation in training data sets. Our work investigates this limitation by exploring methods for customizing and evaluating SLMs via prompting and context engineering.

Our preliminary findings suggest that model family and instruction-tuning approach are at least as important as parameter count for tutoring quality in our context. A 30B coding-specialized model scored near the bottom of our sample, while a general-purpose 4B model in the same family as our top performer scored in the upper half. These results make a practical case for pursuing custom benchmarks when it comes to building LLMs for teaching and learning. Variability across model families is large enough that taking time to design a benchmark with task-relevant criteria is well worth the effort. Generic and general-purpose leaderboards do not predict which models will be the best tutors.

We see several potential directions for next steps. Most pressing is extending the benchmark to handle multi-turn dialogues that reflect actual tutoring sessions more realistically. We attempted to model this in our items, but this part of the test set needs significant expansion. Schroeder et al.~\cite{schroeder2025benchmarks} have argued that single-turn evaluations fail to capture critical problematic behaviors such as sycophancy and abandonment of extended pedagogical tactics. The evaluation set also needs to grow in breadth: with only 17 questions and type-specific criteria scored on subsets as small as one or two, the current benchmark lacks the scale needed for robust per-criterion comparisons. Future iterations should also include held-out test items so that prompt revisions can be evaluated without the confound of tuning on the same data. On the evaluation side, we see the need to validate a reliable SLM-based judge so that multiple trials per model can be scored efficiently, enabling variance estimates and confidence intervals that the current hybrid review process is too labor-intensive to support at scale. Finally, we have an interest in illuminating the boundary conditions on prompt-only approaches to LLM customization, and evaluating whether and to what extent fine-tuned models may outperform the best prompted models.

Our benchmark dataset, rubric, and full evaluation pipeline are available at https://github.com/InviteInstitute/CSTutorBench.

\begin{acknowledgments}
This material is based upon work supported by the National Science Foundation and the Institute of Education Sciences under Award No. 2229612.
\end{acknowledgments}

\section*{Declaration on Generative AI}
During the preparation of this work, the authors used Claude Opus 4.6 in order to support analysis of results and refine/improve grammar and completeness. The authors reviewed and edited the content as needed and take full responsibility for the publication's content.

\bibliography{CSTutorBench_corrected}

\end{document}